%% file: main.tex
\documentclass[10pt,twocolumn,letterpaper]{article}

\usepackage{iccv}              %

\input{preamble}
\definecolor{iccvblue}{rgb}{0.21,0.49,0.74}
\usepackage[pagebackref,breaklinks,colorlinks,allcolors=iccvblue]{hyperref}
\usepackage{xspace}
\usepackage{listings}
\usepackage{minted}
\usepackage[accsupp]{axessibility}  %

\def\paperID{14326} %
\def\confName{ICCV}
\def\confYear{2025}

\title{Teleportraits: Training-Free People Insertion into Any Scene}

\author{Jialu Gao\\
Carnegie Mellon University \\
{\tt\small jialug@andrew.cmu.edu}
\and
K J Joseph\\
Adobe Research\\
{\tt\small josephkj@adobe.com}
\and
Fernando De La Torre\\
Carnegie Mellon University\\
{\tt\small ftorre@cs.cmu.edu}
}

\newcommand{\name}{Teleportraits\xspace}

\begin{document}
\maketitle

\input{sec/0_abstract}
\input{sec/1_intro}
\input{sec/2_related_work}
\input{sec/3_preliminary}

\input{sec/4_method}
\input{sec/5_experiments}
\input{sec/6_conclusion}

{
    \small
    \bibliographystyle{ieeenat_fullname}
    \bibliography{main}
}

\input{sec/Appendix}

\end{document}

%% file: sec/0_abstract.tex
\begin{abstract}
The task of realistically inserting a human from a reference image into a background scene is highly challenging, requiring the model to (1) determine the correct location and poses of the person and (2) perform high-quality personalization conditioned on the background. Previous approaches often treat them as separate problems, overlooking their interconnections, and typically rely on training to achieve high performance. In this work, we introduce a unified \textbf{training-free} pipeline that leverages pre-trained text-to-image diffusion models. We show that diffusion models inherently possess the knowledge to place people in complex scenes without requiring task-specific training. By combining inversion techniques with classifier-free guidance, our method achieves affordance-aware global editing, seamlessly inserting people into scenes. Furthermore, our proposed mask-guided self-attention mechanism ensures high-quality personalization, preserving the subject’s identity, clothing, and body features from just a single reference image. To the best of our knowledge, we are the first to perform realistic human insertions into scenes in a training-free manner and achieve state-of-the-art results in diverse composite scene images with excellent identity preservation in backgrounds and subjects.

\end{abstract}

%% file: sec/1_intro.tex
\section{Introduction}
\label{sec:intro}

Human-centric personalized image content creation has received a lot of attention in recent years due to the increasing demand in commerce for customized experiences, including e-commerce advertising~\cite{du2024towards,wang2023generate}, avatar creation~\cite{instantid, facestudio}, and virtual try-on~\cite{xu2024ootdiffusionoutfittingfusionbased,shen2024imagdressing, zhang2024stablehairrealworldhairtransfer, choi2024improving}. Recent advances in large-scale text-to-image diffusion models~\cite{song2022ddim, ho2020denoisingdiffusionprobabilisticmodels, rombach2022highresolutionimagesynthesislatent, nichol2021glide, dhariwal2021diffusion, saharia2022photorealistic} have given rise to many customization methods that can generate images with the same individual in different scenes, poses, and styles. In this work, we study the problem of personalized human insertion into any scene, or ``person teleportation'': Given a scene image and a human reference image, how can we perform personalized human insertions into the background? 
\begin{figure}[t]
  \centering
  \includegraphics[width=\linewidth]{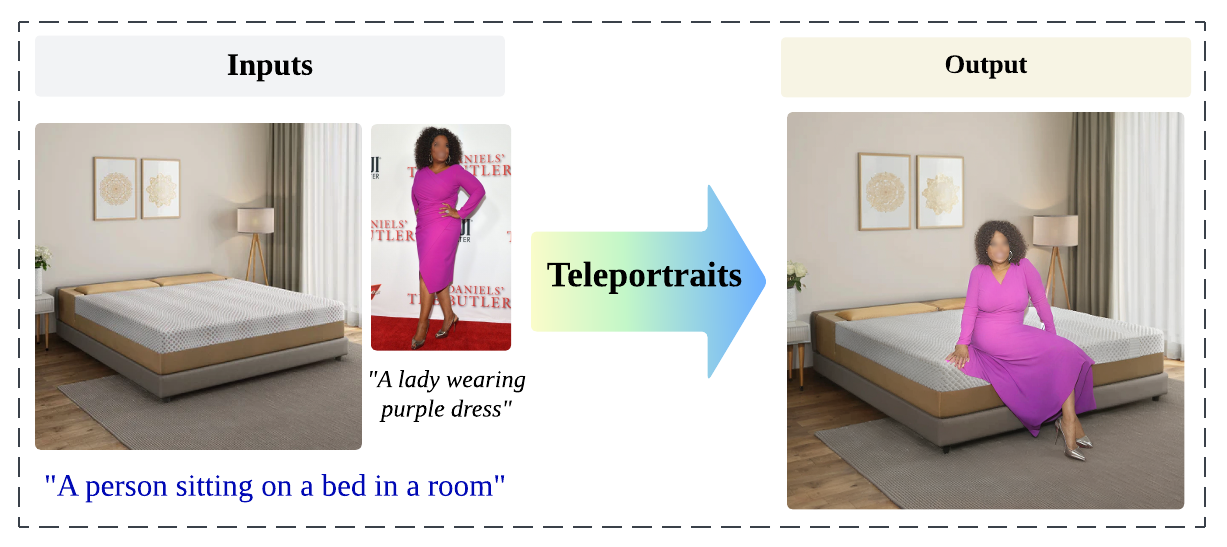}
  \caption{\textbf{Illustration of \name}. \name can insert humans into scenes, while maintaining high degree of affordance.} 
  \label{fig:thumbnail}
\end{figure}

There are two major challenges to this problem: insertion and personalization. The first challenge is highly associated with the concept of affordances, proposed by J.J. Gibson~\cite{affordance} to describe the functional visual relationship between subjects and scenes. To seamlessly insert human subjects into a given background, global affordances reasoning determines the optimal placement, while local affordance understanding refines the subject's precise pose and action. Previous work on global affordances includes human pose and action estimation conditioned on input scenes~\cite{wang2018bingewatchingscalingaffordance, wang2021synthesizinglongterm3dhuman, PuttinghumansinaScene, cao2020longtermhumanmotionprediction}. However, these methods typically rely on training with smaller, curated datasets containing ground-truth annotations, which limits their performance in diverse real-world scenes. Another line of research focuses on local affordances~\cite{SceneAwareHumanPoseGeneration, puttingpeopleplace, anydoor}, aiming to seamlessly synthesize subjects within a given background based on a user-specified location. Nevertheless, such location information is not always available, and absence of global understanding can significantly constrain the effectiveness of local affordance reasoning. Most recently, Text2Place~\cite{t2p} is the first work to consider both levels of affordances, proposing the use of Score Distillation Sampling loss~\cite{dreamfusion} to optimize a human mask parameterized by Gaussian blobs. An off-the-shelf inpainting model~\cite{podell2023sdxl} is then employed to generate human poses at the predicted mask location. Still, the two levels of affordances are treated as separate problems, and test-time tuning is required for each individual scene.

The second challenge is achieving personalization in affordance-aware insertion, generating realistic poses for the person while preserving their facial and clothing features. Current personalization methods can be classified based on whether per-subject optimization is required. Methods such as Textual Inversion~\cite{textual_inversion}, DreamBooth~\cite{ruiz2023dreambooth}, and LoRA~\cite{lora} fine-tune the generation model on reference images to capture visual features. Other methods avoid inference-time tuning by training a lightweight adapter on paired datasets to learn how to extract visual features directly from reference images~\cite{instantid, ip-adapter, arc2face}. However, most personalized generation techniques are conditioned solely on textual inputs. With the introduction of ControlNet~\cite{zhang2023addingconditionalcontroltexttoimage}, structural controls like depth maps and scribbles can also guide generation, yet no prior work has explored conditional personalization using input background images. While, in theory, text-to-image personalization methods could be combined with an inpainting model to enable such conditional personalization, as demonstrated in Text2Place~\cite{t2p}, the quality remains limited since these personalization methods are not specifically trained for inpainting tasks.

In this paper, we propose a unified framework, termed \textbf{\name}, that addresses both challenges simultaneously, in a training free manner. We demonstrate that current large-scale text-to-image diffusion models inherently possess the semantic knowledge required for affordance-aware human insertion. Furthermore, the internal representations of these models can effectively capture and transfer the visual features of the subject for personalization, eliminating the need for additional training or test-time tuning.

Our proposed pipeline operates in three steps. Given a reference subject image and a background scene image, we first approximate the initial noise latents that can reconstruct the two images using inversion techniques. Then, starting from the background noise latent, we apply classifier-free guidance~\cite{ho2022classifierfreediffusionguidance} to direct the model in generating a human within the background using text prompts, for example, \textit{``a person running on the curved road''}. Finally, by leveraging the reference noise latent, we extract internal feature representations from the diffusion model during the self-attention layers.  This allows the generated images to attend to the subject patch through our mask-guided self-attention mechanism, ensuring identity preservation between the subject in the provided reference image and the final generated output.

To demonstrate \name's ability to perform affordance-aware human insertion into diverse scenes, we first evaluate \name on the dataset proposed in Text2Place~\cite{t2p} alongside prior methods. The results show that \name outperforms existing approaches in semantically meaningful human insertion with perfect background preservation. Moreover, we introduce new metrics to assess personalization quality. The quantitative and qualitative results demonstrate that \name excels in generating humans with both global and local affordances, while effectively preserving the subject identity, including facial features, clothing, and body shapes.

In summary, our contributions are as follows: First, we propose a training-free method capable of inserting any person into any scene. Second, we extend the Semantic Human Placement setting introduced in Text2Place to incorporate full-body personalization and introduce new metrics for its evaluation.

\begin{figure*}[t]
  \centering
  \includegraphics[width=0.95\linewidth]{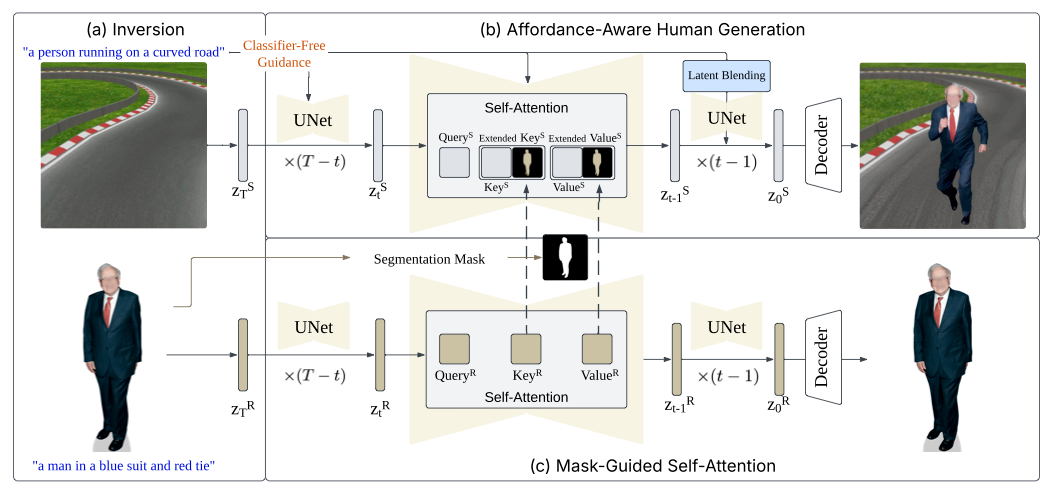}
  \caption{\textbf{Method overview.} \name consists of three steps: (a) Inversion, where we invert the input scene image and reference image into initial latent noise $z_T^S$ and $z_T^R$. This allows \name to utilize the inherent semantic knowledge of diffusion models to place humans and use the hidden representation of diffusion models to perform personalization. (b) Affordance-Aware Human Generation. Starting from the inverted latent $z_T^S$ of the scene image, \name uses classifier-free guidance to gradually guide the model to generate a human at reasonable locations with realistic poses following the text prompt. Latent blending is applied at later denoising steps to ensure background fidelity. (c) Mask-Guided Self-Attention. \name achieves personalization through an extended self-attention mechanism that additionally attend to the keys and values extracted from the recovered diffusion process that reconstruct the reference image.}
  \label{fig:main figure}
\end{figure*}

%% file: sec/2_related_work.tex
\section{Related Work}
\subsection{Subject-driven Image Generation} Subject-driven text-to-image generation focuses on preserving the visual features of any given object or person while performing text-conditioned generation. One approach to capturing a subject's visual characteristics is per-subject optimization, which typically involves learning a special token for each subject~\cite{textual_inversion, DCI_ICO} and/or optimizing the network parameters~\cite{ruiz2023dreambooth, celeb-basis, custom-diffusion}. However, these methods are computationally expensive, as they require optimization for each subject, and they often suffer from overfitting to the reference images. To address these limitations, test-time tuning-free methods have been explored. These approaches leverage pre-trained image encoders to extract visual features and fine-tune the model to condition generation on these extracted features, enabling identity-preserving subject synthesis~\cite{instantid, ip-adapter, elite, blip-diffusion, fastcomposer, photomaker, facestudio, arc2face, masterweaver, stableidentity, subjectdiffusion, uniportrait}. Another direction to solve this problem is through training-free approaches. Consistory~\cite{consistory} extends self-attention by copying keys and values from the reference image and employs DIFT~\cite{DiFT} feature injection to maintain subject consistency in story generation. DreamMatcher~\cite{dreammatcher} proposes Appearance Matching Self-Attention, which warps value patches from the reference image’s self-attention layers using a correspondence mapping to ensure accurate appearance alignment. MagicFace~\cite{magicface} focuses specifically on facial identity preservation, utilizing extended self-attention with keys and values from reference images, guided by aggregated semantic masks. While these methods enable personalization, they are limited to text-conditioned generation and cannot incorporate background images as additional conditioning for personalization.

\subsection{Affordance-aware Subject Insertion} The concept of affordance, introduced by J.J. Gibson~\cite{affordance}, has inspired extensive research in understanding scene-human affordances~\cite{cao2020longtermhumanmotionprediction, wang2018bingewatchingscalingaffordance, PuttinghumansinaScene, wang2021synthesizinglongterm3dhuman}. In image generation, studies on local affordances focus on seamlessly blending a subject into a background at a user-specified location. Anydoor~\cite{anydoor} constructs a dataset of objects and backgrounds with location annotations from videos and fine-tunes a diffusion model to place objects at designated positions. Similarly, Kulal et al.~\cite{puttingpeopleplace} build a dataset of humans and backgrounds from videos, training a model to insert people into predefined locations. While these methods successfully enable affordance-aware local subject insertion, they heavily depend on accurate insertion cues and curated datasets, limiting their effectiveness in diverse real-world scenes where such cues may not always be available. Studies on global affordances focus on estimating plausible human placement within a scene. Wang et al.~\cite{wang2018bingewatchingscalingaffordance} estimate plausible human poses given scenes by learning on extracted poses from sitcom videos. SmartMask~\cite{singh2023smartmaskcontextawarehighfidelity} trains a diffusion model to predict fine-grained mask for subject insertion. Text2Place~\cite{t2p} leverages Score Distillation Sampling loss from DreamFusion~\cite{dreamfusion} to optimize a subject mask parameterized by Gaussian blobs, with an off-the-shelf inpainting model used to generate the person within the mask. Although these methods can perform global affordance-aware subject insertion, they require either expensive test-time tuning for each scene, or rely on large-scale training which limits their ability to generalize to novel scenes. Moreover, they treat localization and human generation as separate problems, overlooking their interdependence, which can degrade the overall quality of the generated image. In contrast, \name unifies the two challenges and achieves training-free human insertion by leveraging the semantic understanding inherently captured by large-scale diffusion models.

%% file: sec/3_preliminary.tex
\section{Preliminaries}
\label{sec:3-preliminary}
\subsection{Latent Diffusion Models}
\label{sec:3-1 ldm}
\begin{figure}[t]
  \centering
  \includegraphics[width=0.45\textwidth]{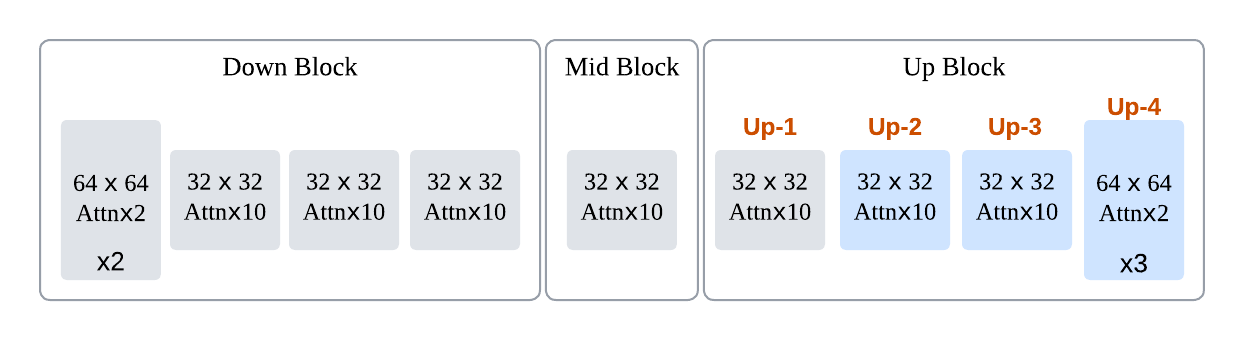}
  \caption{\textbf{SDXL architecture illustration.} SDXL consists of 70 attention layers, where each attention layer includes a cross-attention layer and self-attention layer. In \name, we apply self-attention based personalization on the up-2, up-3, and up-4 layers, as they determine the color, style, and texture details.}
  \label{fig:sdxl}
\end{figure}

Latent Diffusion Models~\cite{rombach2022highresolutionimagesynthesislatent} are a family of diffusion models that use an autoencoder to project images onto the latent space and apply the standard diffusion process~\cite{ho2020denoisingdiffusionprobabilisticmodels, song2022ddim}. The process begins with an initial Gaussian noise $z_T$ and and undergoes a series of denoising steps, where at each timestep $t$, the model predicts $\epsilon(z_t, t)$ that will be used to compute $z_{t-1}$ from $z_t$. At the final timestep $t=0$, the model produces $z_0$, the final sampled image in latent space. In this work, we utilize the publicly available Stable Diffusion XL~(SDXL) model for generation, where the model architecture is illustrated in Fig.~\ref{fig:sdxl}.

\subsection{Self-attention in Diffusion Models}
\label{sec:3-2 sa}
Self-attention mechanism plays a crucial role in maintaining consistency in the style and structure of generation~\cite{dreammatcher, masactrl}. A self-attention layer takes in a hidden feature ${x_{in}} \in \mathcal{R}^{d\times W\times H}$, where $d$ is the feature dimension, and $W$ and $H$ are the resolution of the current attention layer, either 32 or 64 in SDXL model. The hidden feature $x_{in}$ is then mapped to three intermediate representations denoted as Query $\mathbf{Q}$, Key $\mathbf{K}$, and Value $\mathbf{V}$ using linear projections $\mathcal{W}^Q$, $\mathcal{W}^K$, and $\mathcal{W}^V$. The self-attention operation is defined as follows:
$$x_{out} = softmax(\frac{Q^TK}{\sqrt{d}})V$$

\subsection{Classifier-free Guidance}
Classifier-free guidance is proposed by Ho et al.~\cite{ho2022classifierfreediffusionguidance} to trade off controllability with sample fidelity. It samples the noise prediction twice, with and without the conditional text embeddings $c$, and amplifies the difference between them using a guidance weight $w$:
$$\hat{\epsilon}=\epsilon_\theta(x_t, t) + w \cdot (  \epsilon_\theta(x_t, t, c) - \epsilon_\theta(x_t, t))$$

%% file: sec/4_method.tex
\section{Method}
In this section, we present the key components of \name. Given an input scene image $I^{S}$ and a generation prompt $P^S$ describing the scene with human inserted, along with a reference image $I^{R}$ and its corresponding text description $P^R$, \name aims to generate a human inside $I^S$ that both aligns with the scene context described by $P^S$ and resembles the person depicted in $I^R$. 

\name consists of three steps, as shown in Fig.~\ref{fig:main figure}. \name first inverts the scene image $I^S$ and the reference image $I^R$ to obtain obtain their initial latent noise representations, $z_T^S$ and $z_T^R$, respectively~(Sec.~\ref{sec:4-1}). Then starting from $z_T^S$, \name leverages classifier-free guidance to direct the model in generating a human within the scene in accordance with the text prompt $P^s$~(Sec~\ref{sec:4-2}). In the final step, \name introduces mask-guided self-attention, which transfers the visual features of the reference subject into the generated human, ensuring identity preservation while maintaining high-quality generation~(Sec~\ref{sec:4-3}).

\subsection{Inversion}
\label{sec:4-1}
The goal of inversion is to recover an estimated diffusion trajectory that can approximately reconstruct the input scene image $I^S$ and the reference image $I^R$. Inspired by recent work in inversion in diffusion models~\cite{song2022ddim, mokady2022nulltextinversioneditingreal, garibi2024renoiserealimageinversion}, we adopt a similar approach to ReNoise-Inversion~\cite{garibi2024renoiserealimageinversion} and employ a fixed-point iteration strategy. Specifically, the images $I^S$ and $I^R$ are first encoded into the latent space, producing $z_0^S$ and $z_0^R$. At each diffusion timestep $t=0, 1, ..., T-1$, we want to estimate $z_{t+1}$ with $z_{t}$. Since diffusion models are trained to predict $z_{t}$ from $z_{t+1}$ and not in the opposite direction, we initialize an estimate $z^{(0)}_{t+1}$ using DDIM inversion. Then, the estimated $z^{(0)}_{t+1}$ is passed through the UNet, which outputs a noise prediction $\epsilon_\theta^{(0)}$. This noise prediction $\epsilon_\theta^{(0)}$ is then used to renoise $z_{t}$, producing an updated estimate $z_{t+1}^{(1)}$. The same process is repeated across multiple iterations until an accurate estimation of the latent noise $z_{t+1}$. After going through all the timesteps, we obtain the estimated initial latent noise $z_T$.

Our experiments suggest that two iterations are sufficient to reconstruct the input images with high fidelity. As in ReNoise-Inversion, we disable classifier-free guidance during inversion to minimize accumulated error. However, unlike ReNoise-Inversion, we omit the noise averaging step in latent estimation to enhance numerical stability.

\subsection{Affordance-aware Human Generation}
\label{sec:4-2}
The main idea of \name is to utilize the inherent semantic understanding in large-scale diffusion models to perform affordance-aware human insertion in a training-free manner. Once we recover the initial latent noise $z_T^S$ for the input scene image, we want to use the text prompt $P^S$ which describes a plausible and reasonable human insertion solution. By encouraging the model to follow $P^S$, we can achieve seamless human insertion into scenes. To this end, we start with the inverted latent noise $z_T^S$ and utilize classifier-free guidance with a higher guidance weight of $w=7.5$ to encourage the model to generate a person according to the prompt.

A problem with directly relying on text guidance to insert human into scenes is the final generated image may deviate from the original scene image, especially in the background area. To achieve high background fidelity during human generation, we adopt latent blending~\cite{Avrahami_2023} to solve this problem. Specifically, we first perform an inference pass with classifier-free guidance to generate a human inside the scene image. While the background may change, the overall structure and layout of the scene remain faithful to the input $I^S$. Therefore, we use off-the-shelf segmentation model such as SAM~\cite{kirillov2023segment} to obtain a foreground mask and a background mask. During the second inference pass, we apply latent blending using the two masks to ensure that the backgrounds are left intact and the human is being generated into the scene image.
 
\subsection{Mask-guided Self-attention}
\label{sec:4-3}
The final part of \name is to perform personalized generation onto the scene image given reference image $I^R$. To achieve this, we propose mask-guided self-attention to use the hidden representation of diffusion models to transfer visual features of the subject, as depicted in Fig.~\ref{fig:main figure}(c).

As introduced in Sec.~\ref{sec:3-2 sa}, at the core of the diffusion model is a UNet which includes 70 self-attention layers. From the latent noise $z_T^R$ that reproduces the reference image, we can perform a forward diffusion pass and extract the keys and values patches at each self-attention layer. These patches, acting as a natural feature representation for the subject, will be used to transfer the visual identity feature of the subject onto the human being generated. During the generation process described in Sec.~\ref{sec:4-2}, we first retrieve the keys and values from the reference generation process. To ensure that these keys and values contain only visual information of the subject and not the background, we utilize segmentation models on the reference image $I^R$ to obtain a subject mask and apply it on the retrieved keys and values. Then, the keys and values are concatenated with the original keys and values during affordance-aware human generation. In this way, the query patches can attend to both the keys and values from current generation and from the reference image, to allow seamless visual feature transfer.

Following previous works~\cite{agarwal2023imageworthmultiplewords, B-LoRA}, we only apply mask-guided self-attention on part of the up blocks, namely Up-2, Up-3, and Up-4 in Fig.~\ref{fig:sdxl}, where the texture, style, and color of the image are being influenced the most. 

The primary distinction between our mask-guided self-attention and the extended self-attention mechanism used in Consistory~\cite{consistory} is that we apply it solely to the conditional branch in classifier-free guidance, whereas Consistory applies it to both the conditional and unconditional branches. We find that applying it to both branches significantly impedes effective identity transfer. Additionally, Consistory requires an extra feature injection process to achieve consistent subject generation, while in \name, mask-guided self-attention alone is sufficient.

%% file: sec/5_experiments.tex
\section{Experiments}
\label{sec:experiments}
\begin{figure*}[th]
  \centering
  \includegraphics[width=\linewidth]{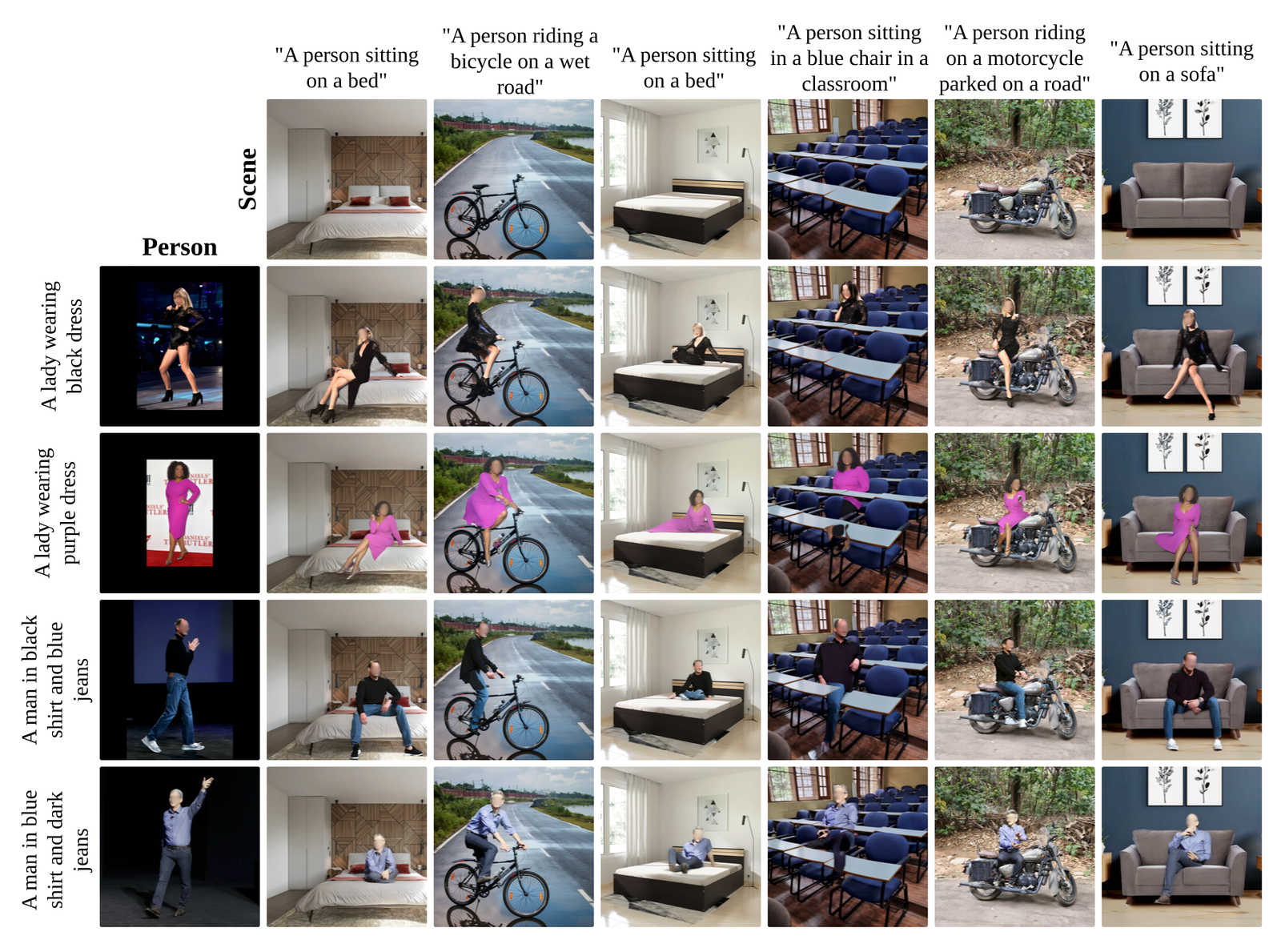}
  \caption{\textbf{Qualitative results.} \name can perform realistic human insertion into various indoor and outdoor scenes given just a single reference image. Results show that \name is able to reason the location and poses of the inserted human, while preserving the subject identity including hair style, clothing, and body shape.}
  \label{fig:qualitative results}
\end{figure*}

We conduct extensive evaluations of \name to assess its ability to realistically insert humans into scenes. Given a scene image and a subject reference image, we prompt BLIP-2~\cite{blip2} to generate a text description of human within the scene, such as ``a person riding a bike on the street''. We also generate a text summary of the subject, like ``a man in a blue suit''. During personalized generation, we replace the ``a person'' in the scene description with the subject’s description to enhance identity preservation. We use SDXL as the default text-to-image model, employing a DDIM sampler with 50 inference steps. Mask-guided self-attention is applied throughout all inference steps, while latent blending is performed during the later denoising timesteps $t\in[10, 20]$ to achieve balance between background preservation and seamless subject integration. All experiments are conducted on a single NVIDIA V100 GPU with 32GB of memory. 

\noindent\textbf{Dataset.} We evaluate \name on the dataset proposed in Text2Place~\cite{t2p}, including 25 celebrities with 26 scenes. During evaluation, we insert each celebrity into all the 26 scenes, resulting in a total of 650 generated images. Notably, the celebrity dataset in Text2Place contains multiple reference images for each person. Since \name only requires one reference image, ideally full-body images as it would provide more information, we select one reference image per person as input for \name. Moreover, to ensure consistency in generation, we resize the reference images so that the human figures are roughly the same size, as shown in the Person column in Fig.~\ref{fig:qualitative results}.

\noindent\textbf{Metrics.} We first follow Text2Place and evaluate \name on these metrics for realistic human placement: \textbf{(1)} CLIP-T, which computes the cosine similarity between the CLIP~\cite{clip} embeddings of the text prompt and the final generated image to measure prompt alignment. \textbf{(2)} Person generation, which uses SAM~\cite{kirillov2023segment} to detect whether a human is being generated in the final image and calculate the percentage of successful human insertion. \textbf{(3)} Background preservation, which uses SAM to mask out the human subject and compute the LPIPS~\cite{zhang2018unreasonableeffectivenessdeepfeatures} score between the generated image and the original scene image to measure background fidelity. Furthermore, we include two metrics to measure identity preservation in personalization following prior works~\cite{ruiz2023dreambooth}: \textbf{(4)} CLIP-I, which calculates the cosine similarity between the CLIP Image embedding of the generated human image and the reference image. \textbf{(5)} DINO, which is the average cosine similarity between the ViTS/16 DINO~\cite{caron2021emerging} embeddings of the generated and reference human image. 

\noindent\textbf{VLM evaluation.} As discussed in
 prior work~\cite{peng2025dreambenchhumanalignedbenchmarkpersonalized}, automated evaluations of personalization models can be misaligned with humans. For a more comprehensive evaluation, we adopt and extend the GPT-based evaluation protocol from~\cite{visualpersona} to assess subject identity preservation~(\textbf{VLM-S}), text alignment~(\textbf{VLM-T}), and background preservation~(\textbf{VLM-BG}). Details of the evaluation pipeline and GPT prompts are provided in the Appendix.~\ref{appendix: vlm-eval}.

\noindent\textbf{Human evaluation.} We perform human evaluation following DreamMatcher~\cite{dreamfusion} with 51 users and 36 samples from the Text2Place dataset. Users rank the generated images according to subject identity, text alignment, and scene consistency. Additional
details can be found in Appendix.~\ref{appendix: human-eval}.

\noindent\textbf{Baselines.} We compare \name with multiple baselines to show its superior performance in both affordance-aware human placement and personalization. We first compare \name with \textbf{Text2Place} on the overall task of personalized human insertion into scenes. Since original Text2Place operates on multiple reference images, we also compare with Text2Place using a single reference image, which we call \textbf{Text2Place~(single)}. Additionally, we compare \name's ability to perform affordance-aware human placement with Text2Place by using the human mask extracted from our generated image as the mask for personalized inpainting in Text2Place, which is denoted as \textbf{Mask+Text2Place}. Lastly, we compare \name with a state-of-the-art, open-source zero-shot object insertion method, \textbf{AnyDoor}~\cite{anydoor}. Since a mask is also required for Anydoor, we use the human mask extracted from our final generated image and measure Anydoor's performance on zero-shot personalized human insertion.

\subsection{Qualitative Results}
We first present the qualitative results of \name in Fig.~\ref{fig:qualitative results}. Our method can achieve realistic human insertion into various scenes following the text prompts, and can reason the correct human poses that interact with the objects in the scenes accordingly, such as riding on the bike or sitting in a chair. Moreover, our method can perform high-quality personalization given only a single reference image, perfectly maintaining the person identity, including hair style, clothing, and body features. 

\begin{figure*}[th]
  \centering
  \includegraphics[width=\linewidth]{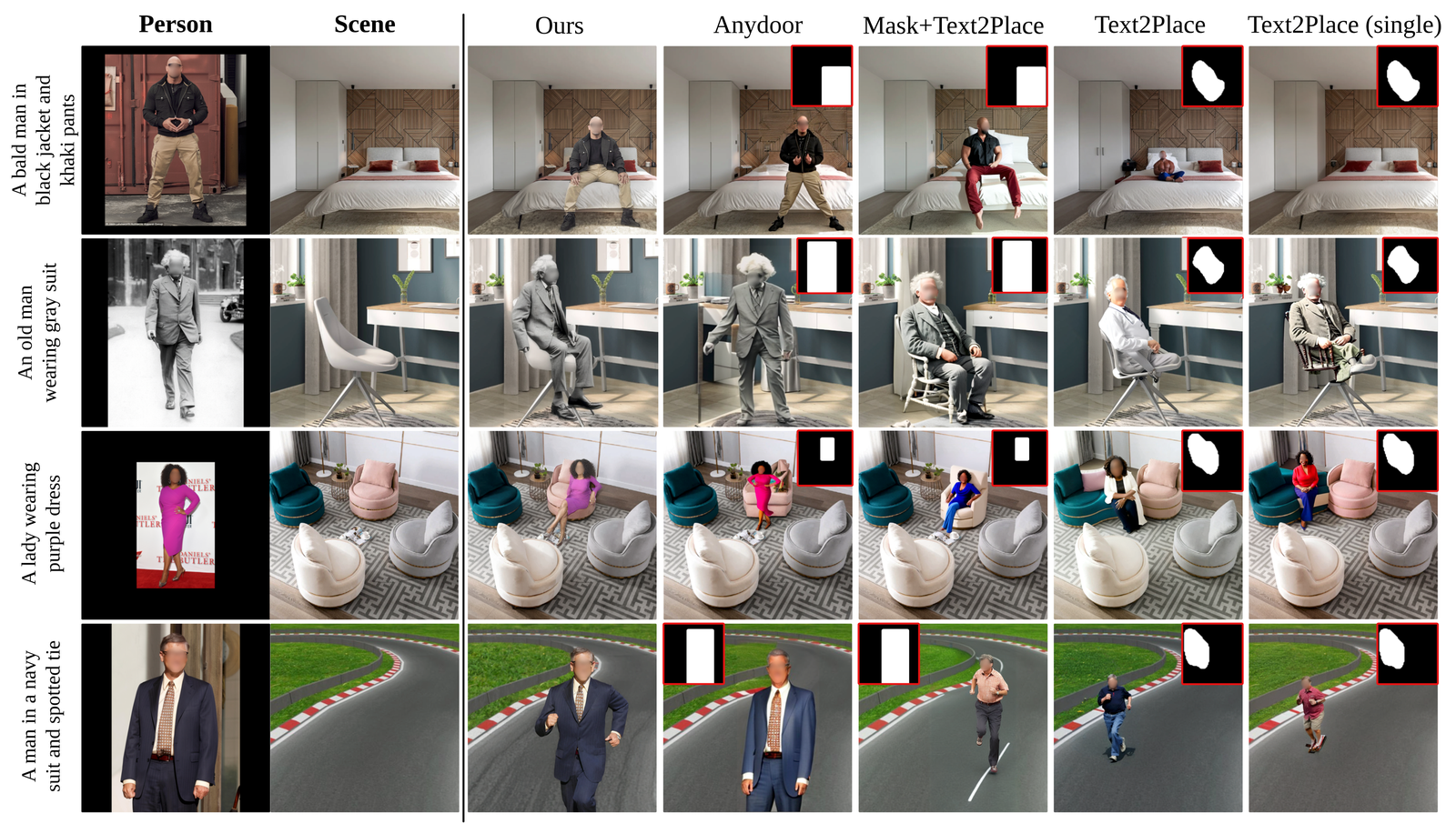}
  \caption{\textbf{Qualitative comparison with baselines.} Using the bounding box from the generated human in \name, Anydoor is able to insert human but fails to generate realistic human poses that interacts with the scene. Compared to Text2Place, \name can not only generate better location for human insertion, leading to better inpainting results, but also can preserve the human identity much better.}
  \label{fig:baseline comparison}
\end{figure*}

In Fig.~\ref{fig:baseline comparison}, we compare \name with the baselines. Firstly, Text2Place sometimes fails to generate the most suitable mask blobs, leading to unsuccessful human insertion. In addition, Text2Place cannot successfully transfer clothing and body shape features, either with multiple reference images or with only a single reference image. In contrast, our method perfectly preserves the clothing details of the subjects without any test-time tuning. In Mask+Text2Place, we enhance Text2Place by using the bounding box of the generated person from \name as the subject mask for Text2Place’s inpainting module. Results show that this leads to more successful inpainting, demonstrating that \name produces more accurate and affordance-aware human insertion. For Anydoor, we provide the extracted human bounding box from \name as input. While Anydoor preserves subject identity well, it fails to generate plausible human poses that meaningfully interact with surroundings. This is likely because Anydoor is trained on large-scale internet videos with objects, making it less effective for human-centric generation.

\subsection{Quantitative Results}
\begin{table*}
  \centering
  \begin{tabular*}{\textwidth}{@{\extracolsep{\fill}}lccccccccc@{}} 
    \toprule
     & CLIP-T$\uparrow$ & Person (\%)$\uparrow$ & LPIPS$\downarrow$ & CLIP-I $\uparrow$ & DINO$\uparrow$ & VLM-S$\uparrow$ & VLM-T$\uparrow$ & VLM-BG$\uparrow$ \\
    \midrule
    Text2Place & 0.267 & 84.2 & 0.094 & 0.573 & 0.164 & 2.25 & 3.85 & \underline{6.17} \\
    Text2Place~(single) & 0.267 & 86.2 & 0.093 & 0.567 & 0.180 & 1.93 & 3.79 & 6.00\\
    Our Mask + Text2Place & 0.269 & 91.8 & 0.075 & 0.572 & 0.177 & 2.18 & \underline{4.17} & 6.02 \\
    Our Mask + AnyDoor & \underline{0.275} & \textbf{100.0} & \underline{0.053} & \textbf{0.681} & \textbf{0.447} & \textbf{4.05} & 1.82 & 5.58\\
    Ours & \textbf{0.287} & \underline{97.4} & \textbf{0.025} & \underline{0.596} & \underline{0.244} & \underline{4.01} & \textbf{4.93} & \textbf{6.29} \\
    \bottomrule
  \end{tabular*}
  \caption{\textbf{Quantitative results.} Compared to Text2Place, \name has better performance across all metrics. For affordance-aware human placement, \name can predict the location better, leading to high success-rate for Text2Place's inpainting pipeline to generate a human. While Anydoor can generate human more similar to the references, it falls short in following the text prompt, failing to generate semantically meaningful poses for the inserted human.}
  \label{tab:baseline}
\end{table*}

\begin{figure}[h]
    \centering
    \includegraphics[width=\linewidth]{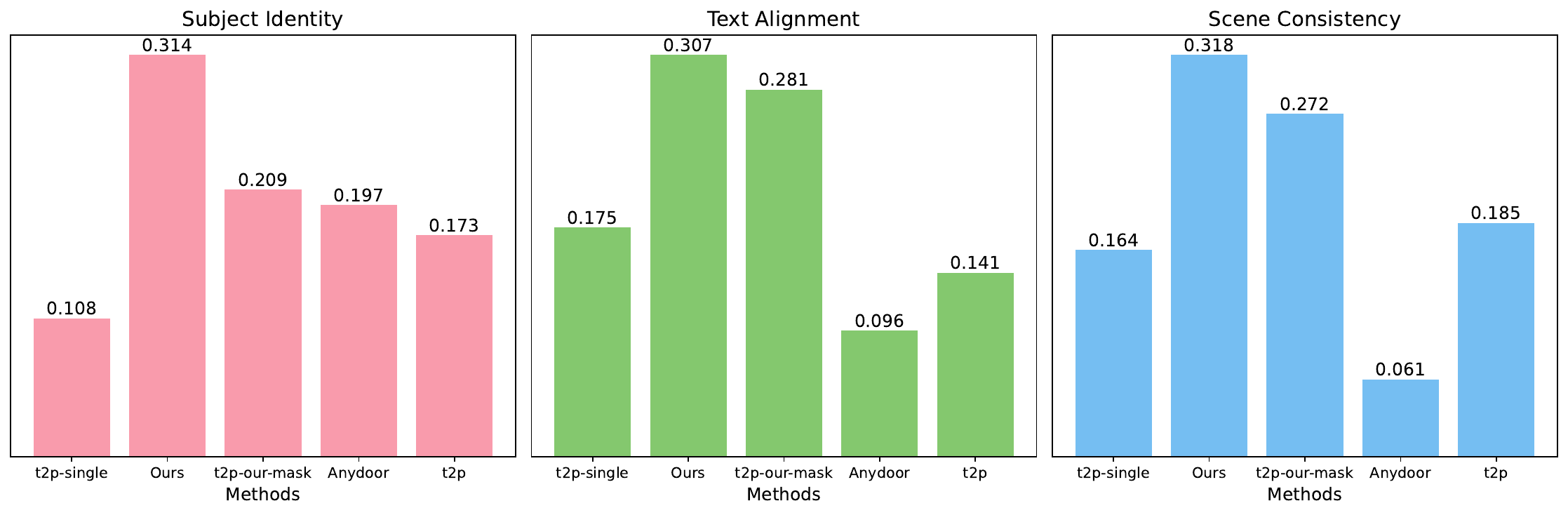}
    \caption{\textbf{Human Evaluation Results.} Following DreamMatcher~\cite{dreammatcher}, 51 users ranked 36 samples from the Text2Place dataset based on subject identity, text alignment, and scene consistency. \name achieves the highest scores across all aspects.}
    \label{fig:human-eval}
\end{figure}

Next we present the quantitative evaluation results on \name and the baselines, with automated evaluation results in Table.~\ref{tab:baseline} and human study results in Figure~\ref{fig:human-eval}. \\
\vspace{-10pt} 

\name shows the best performance in text-to-image alignment and background preservation compared to all baselines, demonstrating its superior ability to semantically insert human into scenes while leaving the background almost unchanged. By replacing the predicted mask in Text2Place with ours, we are able to achieve higher human insertion success rate, demonstrating \name's state-of-the-art ability to accomplish the task of semantic human placement. Additionally, \name achieves a higher success rate in human insertion than Text2Place, suggesting that the pre-trained inpainting model used in Text2Place has worse performance compared to our training-free approach. This highlights the effectiveness of \name in handling human insertion tasks without relying on task-specific training, and emphasizes the advantage of using semantic knowledge within diffusion models to perform global editing. For personalization quality, \name achieves higher identity preservation scores compared to all baselines, except Anydoor. Anydoor’s better performance in identity similarity is due to its tendency to copy-paste the human directly from the reference image into the scene, without adjusting the subject's pose or angle. As shown in Fig.~\ref{fig:baseline comparison}, this results in a higher similarity between the reference image and the generated subject, but limits its ability to produce natural, dynamic human poses.

Overall, both the qualitative and quantitative results demonstrate \name's capability in solving the two challenges in human insertion into scenes: affordance-aware human insertion and conditional personalization given background scenes. Notably, \name achieves the task in a training-free manner, costing less than 1 min to generate a single image, while Text2Place requires per-subject optimization and can take up to 1 hour to generate a novel subject in a novel scene.
\begin{figure}[t]
  \centering
  \includegraphics[width=\linewidth]{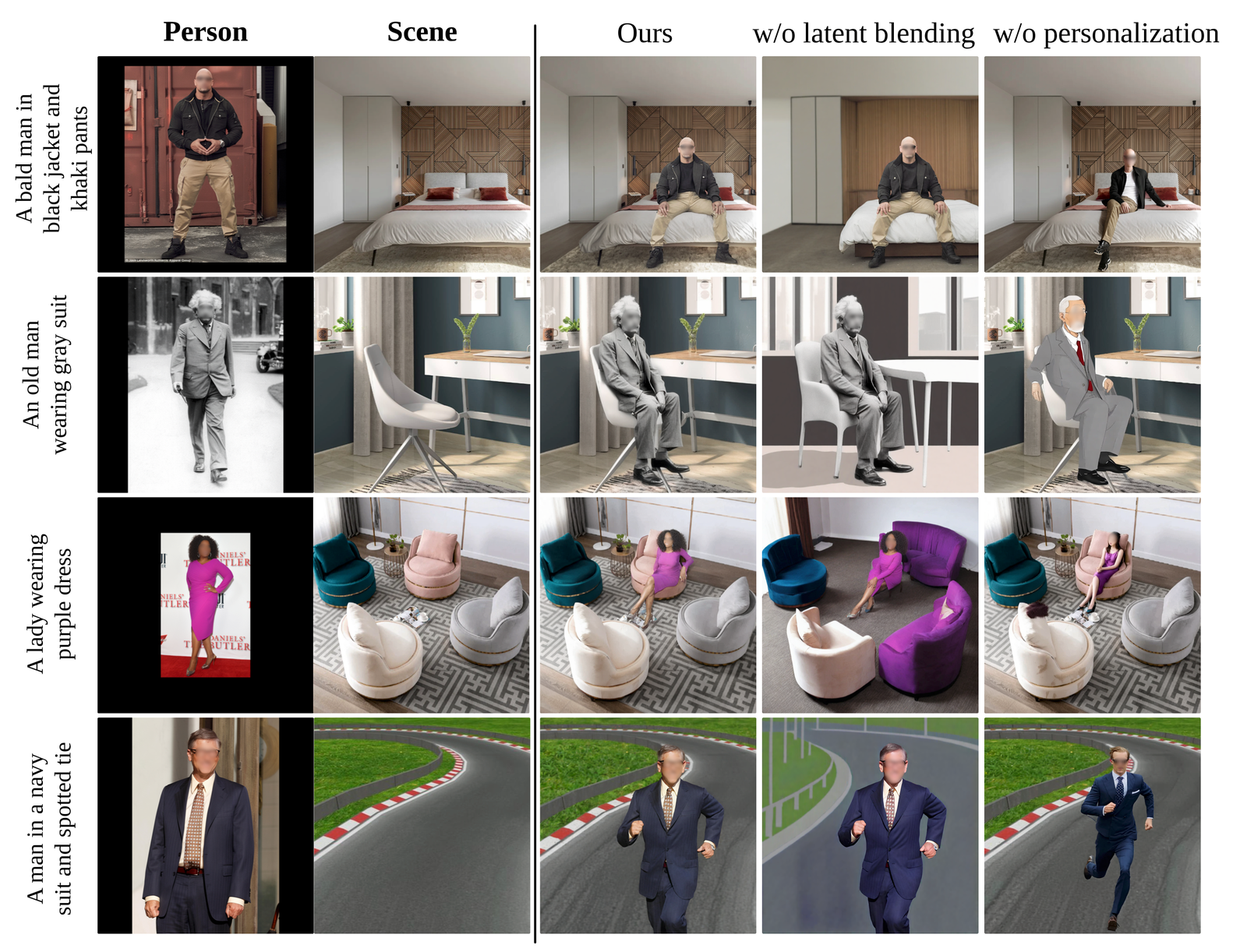}
  \caption{\textbf{Ablation study of \name.} Results show that latent blending plays an important role in background preservation, and our personalization module can successfully transfer detailed visual features of the reference subject into the generated image.}
  \label{fig:ablation}
\end{figure}

\subsection{Ablations}
We now present the ablation study on the following components in \name: (1) Latent blending during semantic human generation (2) Mask guided self-attention for personalization. The qualitative results are displayed in Fig.~\ref{fig:ablation}, and quantitative results are reported in Table.~\ref{tab:ablation}. More ablation results can be found in Appendix.~\ref{appendix:ablation}.

Latent blending plays a crucial role in preserving background fidelity. When removed, the overall structure of the image still resembles the input scene, but the detailed appearance of the background changed. This is because the structure is largely influenced by the initial noise~\cite{add-it}. Therefore, even under the influence of classifier-free guidance, when the generation starts from the same latent noise that reconstructs the scene image, it will automatically follow the scene layout and place humans at reasonable locations. However, the detailed appearance of the image can be largely influenced by the text prompt, thus latent blending is required to preserve the visual details of background. 

Personalization using mask-guided self-attention is crucial to preserving high-level and low-level subject identity. When generating without mask-guided self-attention, we can see that the model generates a human that roughly resembles the reference image, because the subject prompt already contains some high-level information about the appearance, such as the color of the clothing. However, text has limited granularity and thus cannot encode detailed visual features, such as the buttons on the jacket and the color of the spotted tie. This is where mask-guided self-attention comes in and performs detailed visual feature transfer from the reference image to the final generated image.

\begin{table}
  \centering
  \begin{tabular}{p{2cm} p{1.5cm} p{1.5cm} p{2cm}}
    \toprule
     & Ours & w/o Latent blending & w/o Personalization \\
    \midrule
    CLIP-T$\uparrow$ & 0.287 & \underline{0.288} & \textbf{0.289}  \\
    Person(\%)$\uparrow$ & \underline{97.4} & \textbf{98.9} & 94.3  \\
    LPIPS$\downarrow$ & \textbf{0.025} & 0.189 & \underline{0.029}  \\
    CLIP-I $\uparrow$ & \textbf{0.596} & \underline{0.590} & 0.536 \\
    DINO$\uparrow$ &\textbf{0.244} & \underline{0.236} & 0.174  \\
    \bottomrule
  \end{tabular}
  \caption{\textbf{Ablation study on \name.} Quantitative results also demonstrate the role of latent blending in background preservation. And subject similarity metrics show that our personalization method greatly contributes to better identity preservation.}
  \label{tab:ablation}
\end{table}

%% file: sec/6_conclusion.tex
\section{Conclusion}
In this work, we present \name, a training-free method that builds on pre-trained diffusion models and successfully tackles the challenging problem of inserting humans into any scene with a single reference image. \name achieves affordance-aware human insertion by leveraging the semantic knowledge inherent in large-scale diffusion models, while transferring the visual features of the subject through a mask-guided self-attention process. We hope this work sheds light on the potential of training-free approaches that utilize the inherent knowledge of diffusion models to perform various image manipulation tasks.

%% file: sec/Appendix.tex
\clearpage
\setcounter{page}{1}
\maketitlesupplementary

\appendix
\section{Limitation}
\label{sec:limitation}
While \name has demonstrated state-of-the-art performance in the task of human insertion into scenes, there are some limitations to the method. 

\begin{figure}[h]
  \centering
  \includegraphics[width=\linewidth]{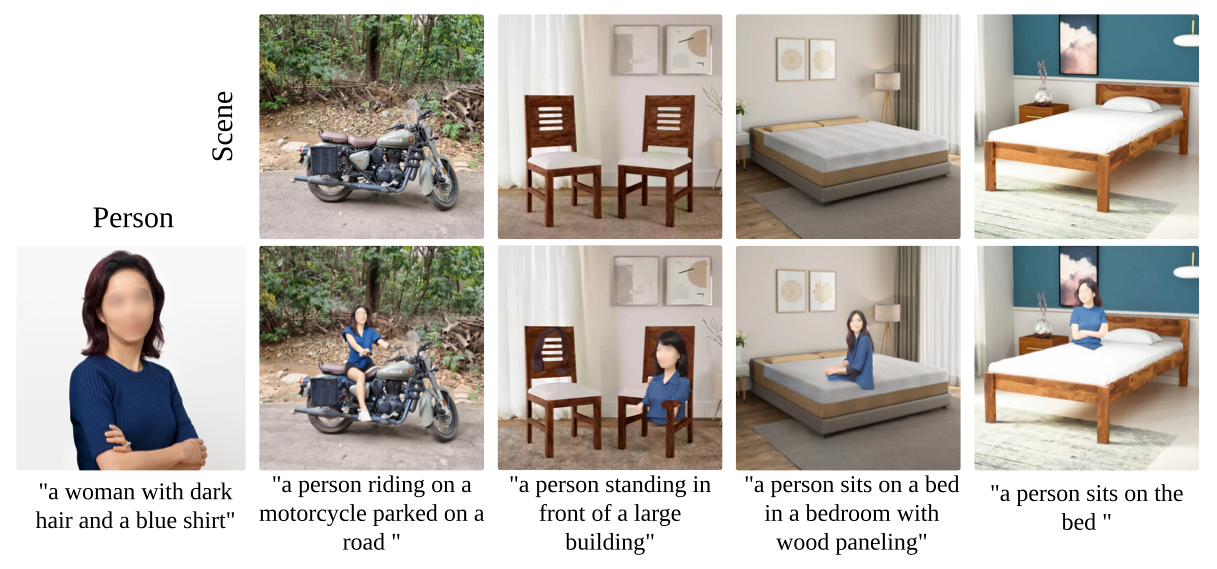}
  \caption{\textbf{Failure case 1.} When the reference image only contains a small part of the body, the personalized generation quality degrades.}
  \label{fig:lim1}
\end{figure}

Firstly, \name performs the best with full-body images as reference, and will suffer from problems like low-quality personalization and disproportional human sizes if the reference image only contains the upper body, or only the face of the human~(Fig.~\ref{fig:lim1})

\begin{figure}[h]
  \centering
  \includegraphics[width=\linewidth]{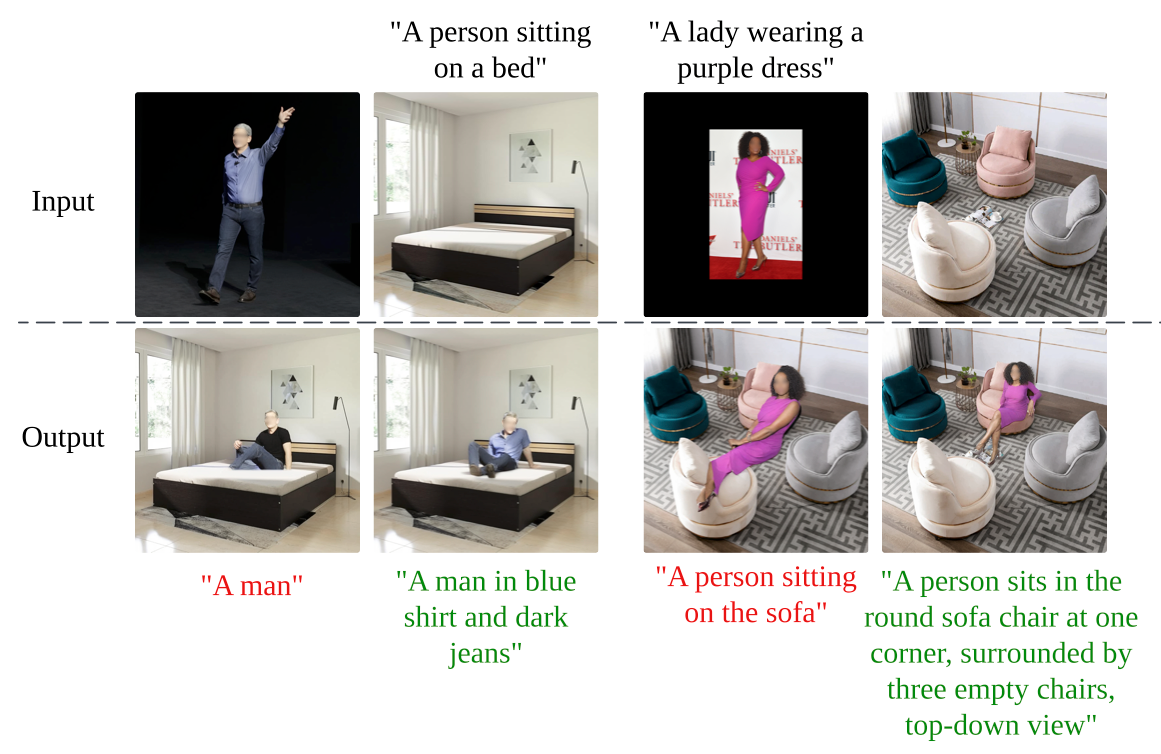}
  \caption{\textbf{Failure case 2.} The influence of text prompts with complex examples.}
  \label{fig:lim2}
\end{figure}

Secondly, the quality of the generation is influenced by the text prompt, especially when the scene is complex or the person has many detailed visual characteristics to capture. For example, a shorter prompt like ``a person sitting on the bed'' will lead to worse result compared to a more detailed prompt like ``a man wearing blue shirt and dark jeans sitting on the bed''. Another example would be ``a person sitting on the sofa'' leads worse result compared to a more detailed prompt like ``a person sits in the round sofa chair at one corner, surrounded by three empty chairs, top-down'', on a scene containing multiple sofas captured from top-down view~(Fig.~\ref{fig:lim2}). This is probably due to the bias in large-scale internet dataset that the diffusion model is trained on, but overall, for common scene images and people, the effort for prompt tuning is minimal.

\section{More Ablation Results}
\label{appendix:ablation}
Here we present more ablation studies on the hyper-parameters used in \name.\\

\begin{figure*}[h]
  \centering
  \includegraphics[width=\linewidth]{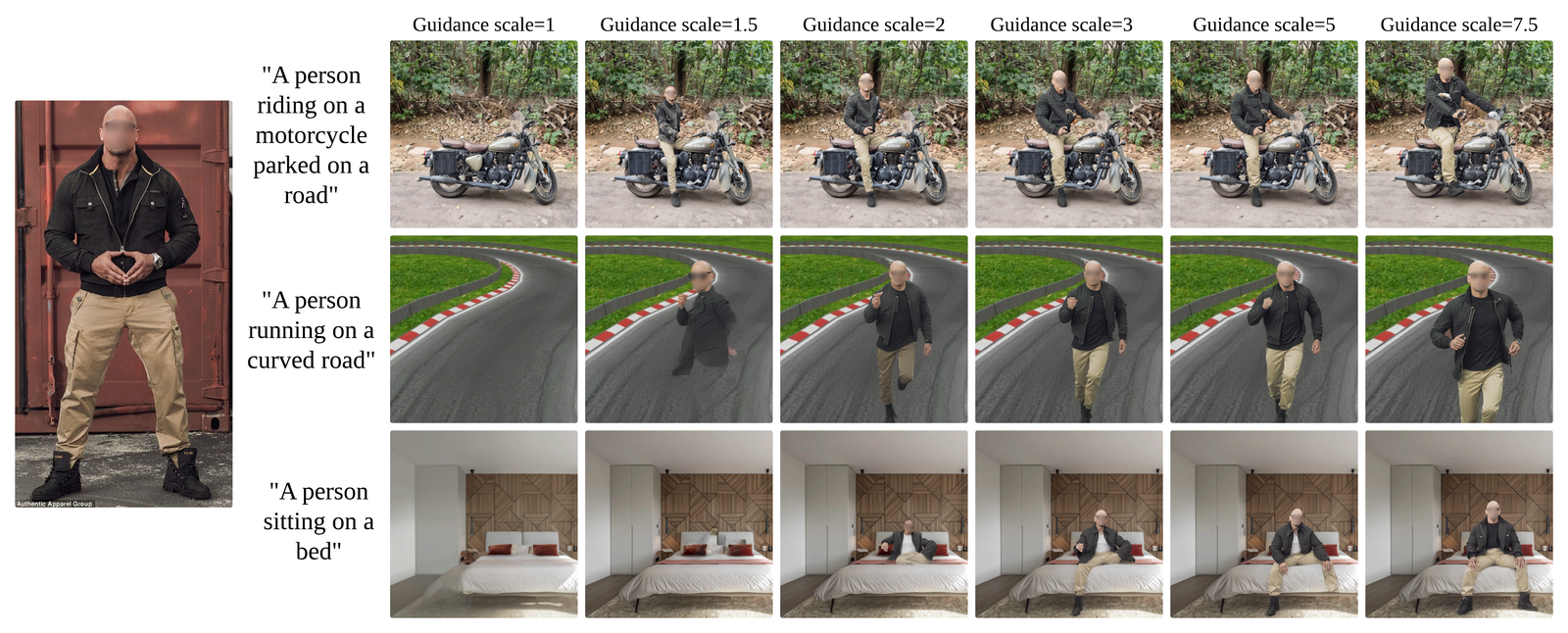}
  \caption{\textbf{Influence of guidance-scale.} Results show that with a larger guidance scale, we can achieve better human insertion into scenes because the generation process will be guided more towards the text prompt, which describes the scene containing a human.}
  \label{fig:ablation-1}
\end{figure*}

\noindent\textbf{Influence of classifier-free guidance scale.} In Fig.~\ref{fig:ablation-1}, we present the effect of different classifier-free guidance scale has on the final generated images. With a guidance scale of $1$, it is equivalent to disabling classifier-free guidance, and therefore only the scene image is reconstructed and no human is being generated. With the guidance scale increasing, we can observe that the human being generated is getting clearer and clearer, taking up more space in the image. This is because a larger guidance scale will drive the generation more towards the direction of text prompt, where a human is included. \\

\begin{figure*}[h]
  \centering
  \includegraphics[width=\linewidth]{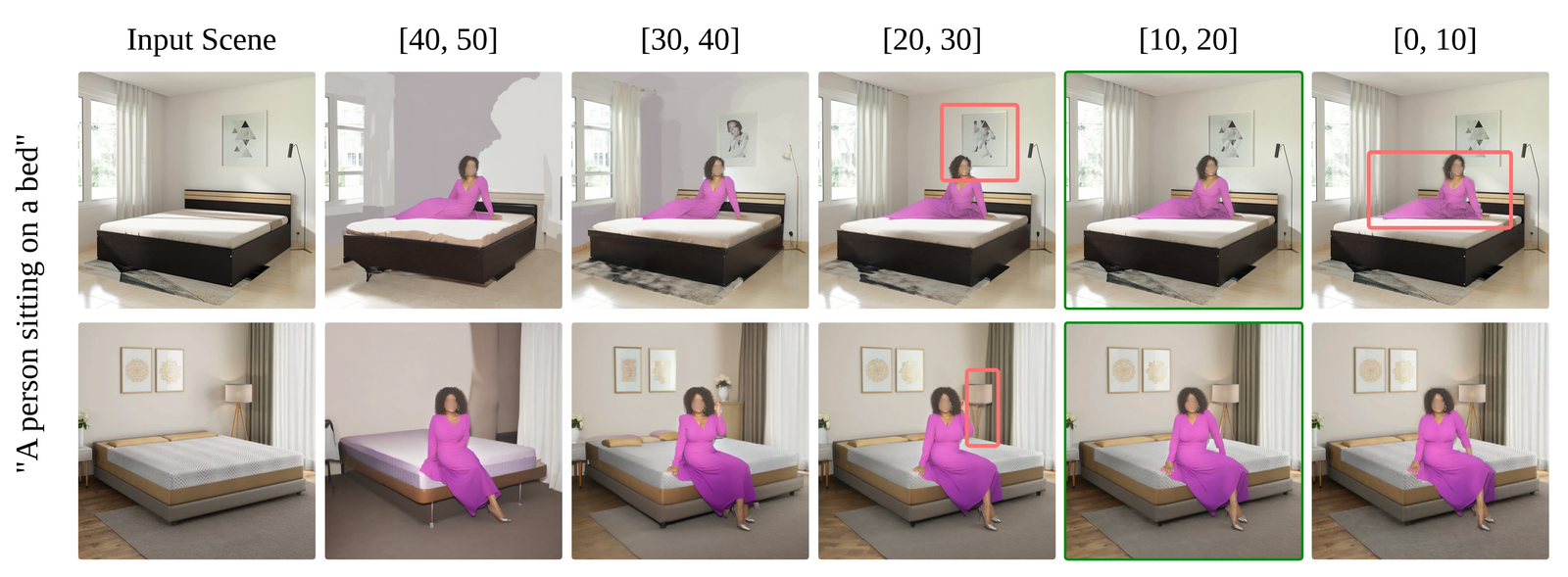}
  \caption{\textbf{Influence of latent blending timesteps}. We report results obtained by applying latent blending during $t \in [0, 10], [10, 20], [20, 30], [30, 40], [40, 50]$. The Denoising process starts from $t=50$ and ends in $t=0$, meaning that larger $t$ indicates earlier diffusion steps, and smaller $t$ represents later steps. Results show that applying latent blending during $t\in [10, 20]$ achieves a perfect balance between background preservation and seamless foreground blending.}
  \label{fig:ablation-2}
\end{figure*}

\noindent\textbf{Influence of latent blending timesteps.} In Fig.~\ref{fig:ablation-2}, we show how different latent blending timesteps influences the output images. We can observe that applying latent blending during earlier timestep results in more obvious changes in backgrounds. This is because diffusion models usually determine the structure and layout during early timesteps, and detailed appearances are determined during the later timesteps. When we move the $t$ range to later timesteps, we can see that the background fidelity significantly increases. However, if we only apply latent blending right before the denoising process finishes, it may result in visual artifacts such as a glow surrounding the subject. Therefore, we choose to apply latent blending during $t \in [10, 20]$ in \name to achieve a balance between background preservation and overall image quality. \\

\begin{figure*}[h]
  \centering
  \includegraphics[width=\linewidth]{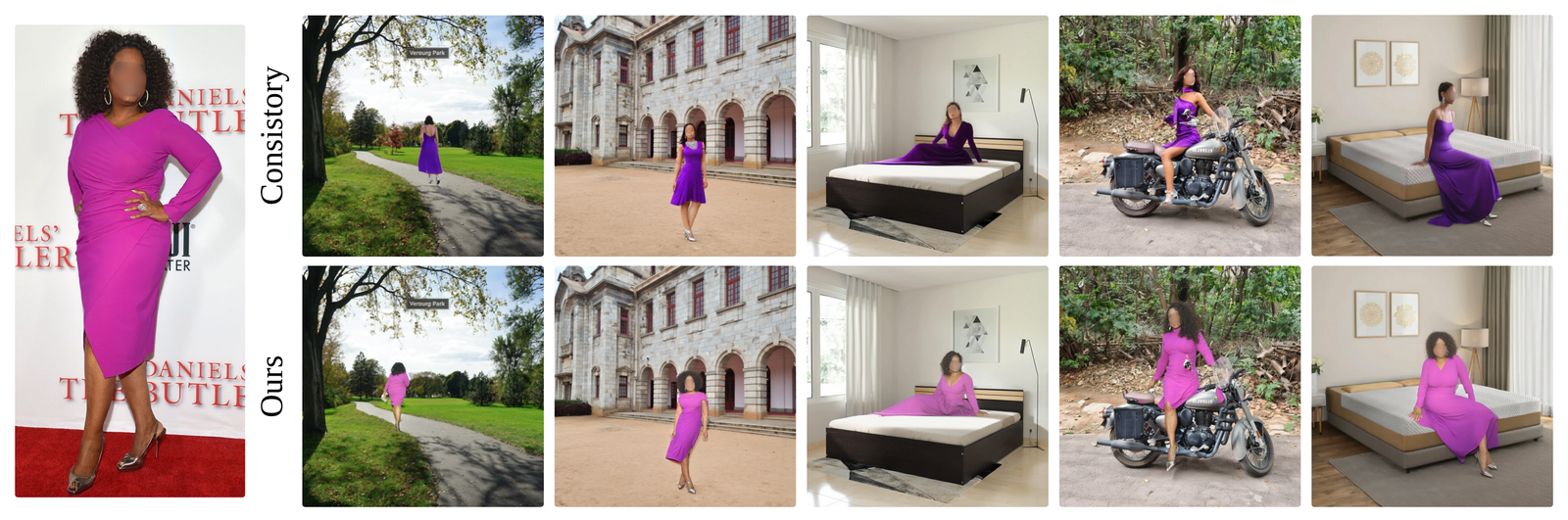}
  \caption{\textbf{Influence of whether applying self-attention feature transfer on the unconditional branch}. Results show that only applying mask-guided self-attention on the conditional branch as in \name can significantly increase the personalization performance, generating subjects highly similar to the reference.}
  \label{fig:ablation-3}
\end{figure*}

\noindent\textbf{Influence of performing mask-guided on the unconditional branch.} Here we compare our mask-guided self-attention mechanism with the one proposed in Consistory~\cite{consistory}. In particular, the main difference between our method and the one used in Consistory is that we are only applying the mask-guided self-attention on the conditional generation branch of classifier-free guidance. In contrast, Consistory applies it on both the conditional branch and unconditional branch during generation. We report the results in Fig.~\ref{fig:ablation-3}, which clearly indicates that applying modified self-attention on both conditional and unconditional branches during generation largely degrades the personalization quality, demonstrating \name's superior performance in transferring visual features from a single reference image into various scenes during human generation.

\section{VLM Evaluation Details}
\label{appendix: vlm-eval}
\begin{figure*}[h]
    \centering
    \includegraphics[width=\linewidth]{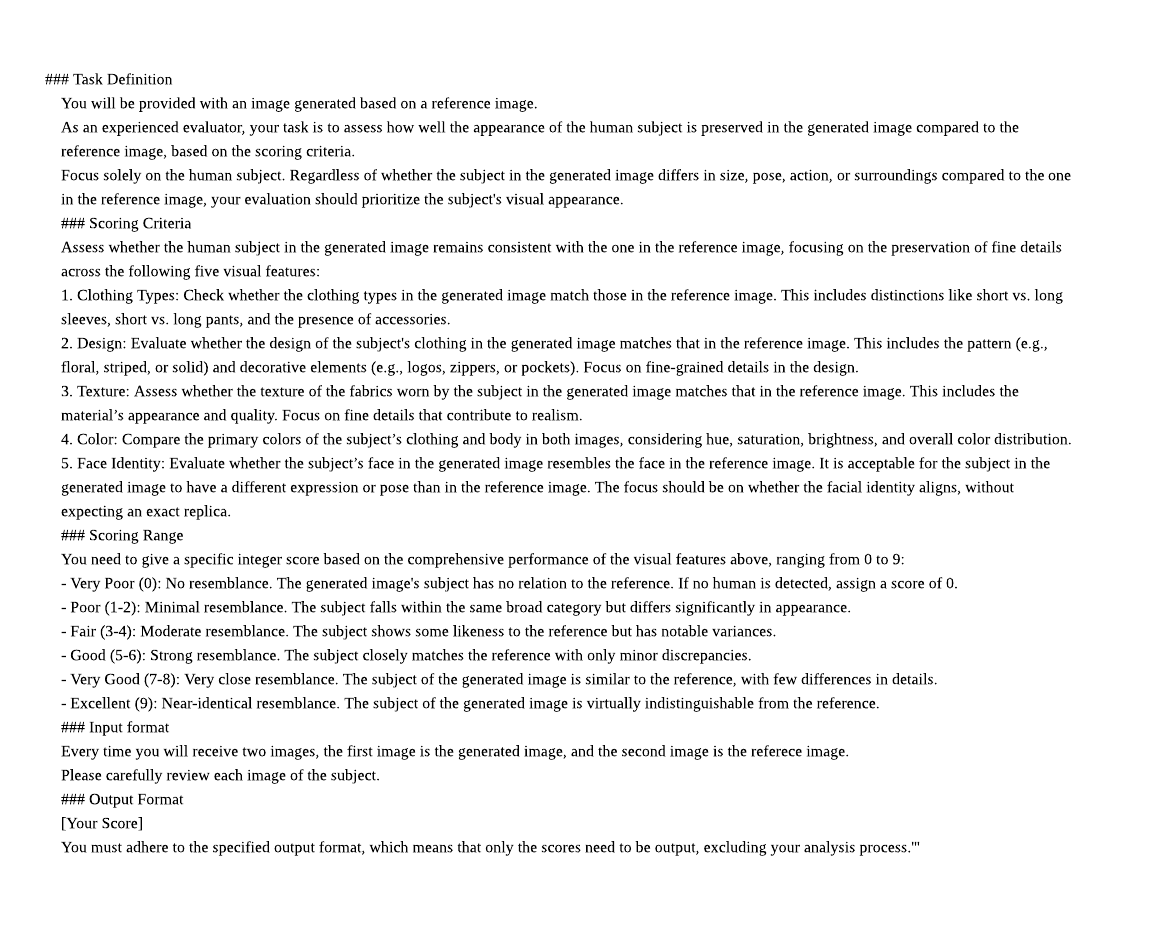}
    \caption{\textbf{GPT prompts for evaluating personalization quality.}}
    \label{fig: vlm-prompt-subject}
\end{figure*}

\begin{figure*}[h]
    \centering
    \includegraphics[width=\linewidth]{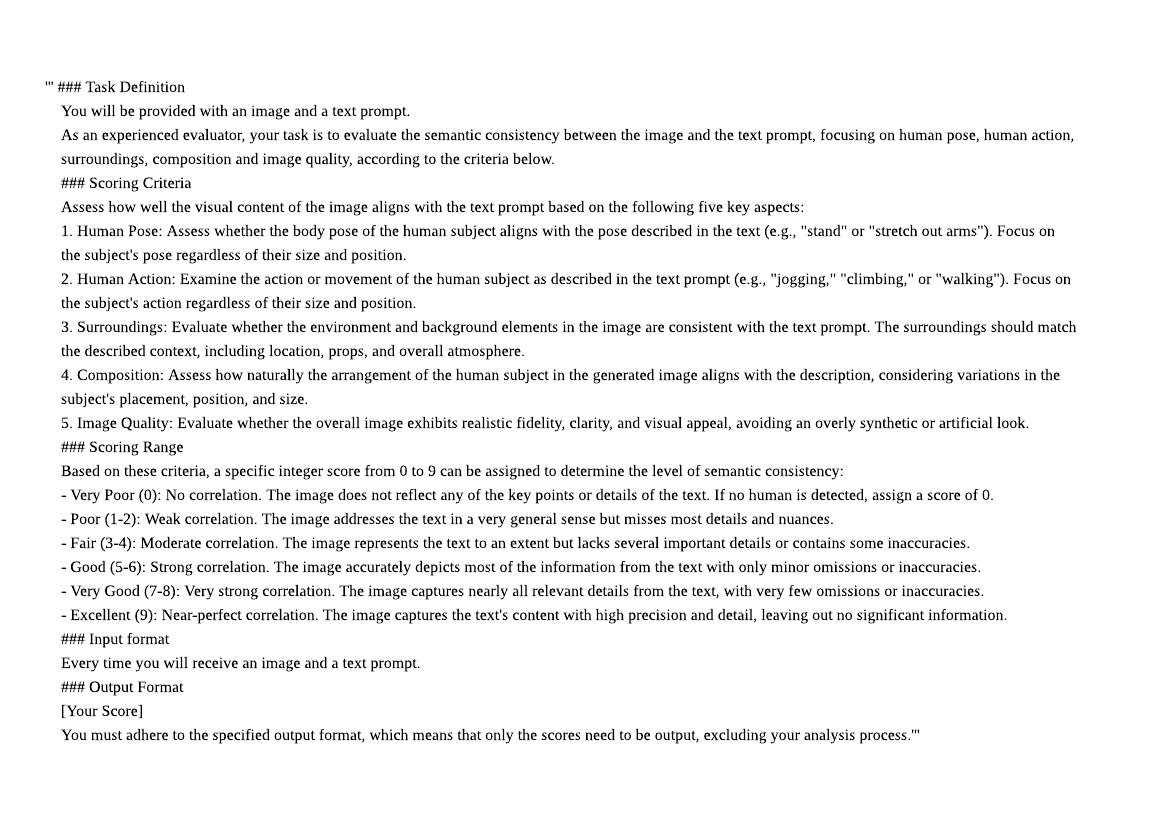}
    \caption{\textbf{GPT prompts for evaluating prompt alignment.}}
    \label{fig: vlm-prompt-text}
\end{figure*}

\begin{figure*}[h]
    \centering
    \includegraphics[width=\linewidth]{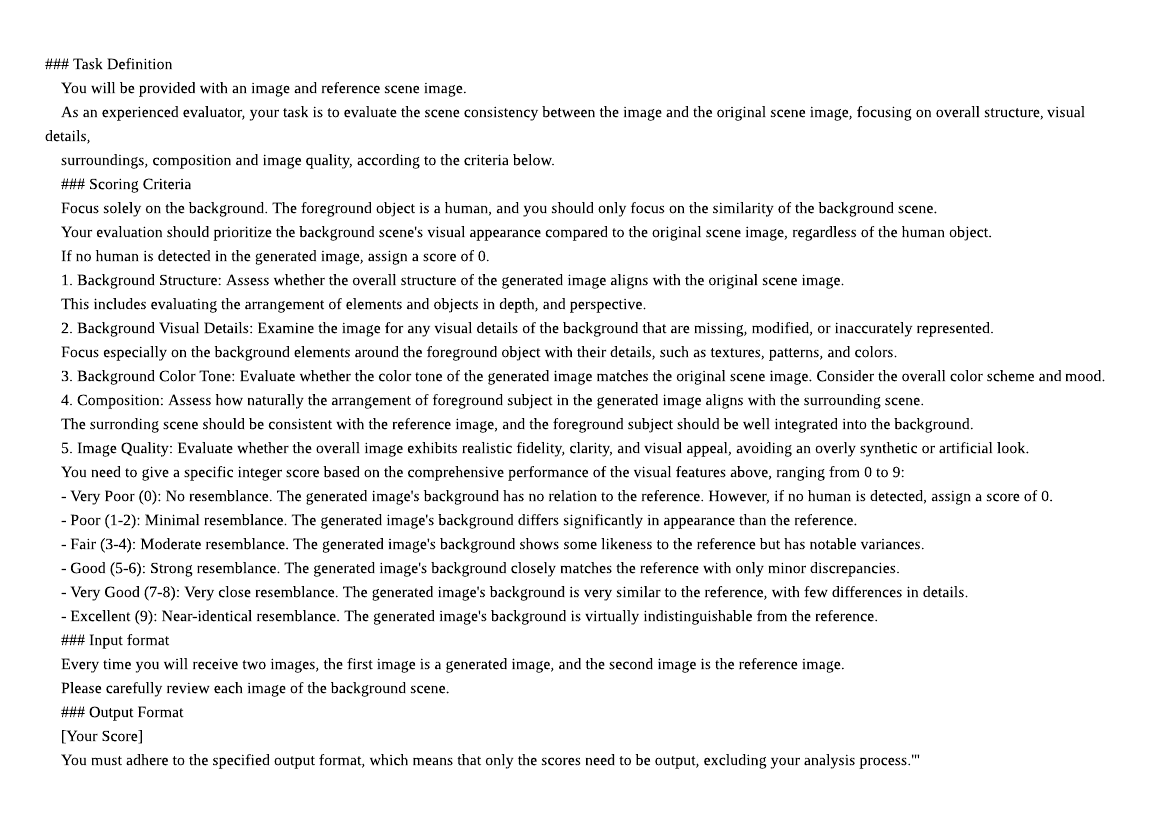}
    \caption{\textbf{GPT prompts for evaluating background fidelity during insertion.}}
    \label{fig: vlm-prompt-background}
\end{figure*}

Following the GPT evaluation protocol in~\cite{visualpersona}, we designed three different prompts for evaluating \name's ability in subject identity preservation~(Fig.~\ref{fig: vlm-prompt-subject}), text alignment~(Fig.~\ref{fig: vlm-prompt-text}), and background scene preservation~(Fig.~\ref{fig: vlm-prompt-background}). The GPT model version is GPT-4o, and all evaluations are performed with a temperature of 0 and high image details.

\section{Human Evaluation Details}
\label{appendix: human-eval}
We conducted a paired human preference
study on subject fidelity, prompt alignment, and background fidelity, comparing \name to the baseline works as listed in Sec.~\ref{sec:experiments} of the main paper. The results are summarized in Fig.~\ref{fig:human-eval} in the main paper. \\

\begin{figure*}[h]
    \centering
    \includegraphics[width=0.75\linewidth]{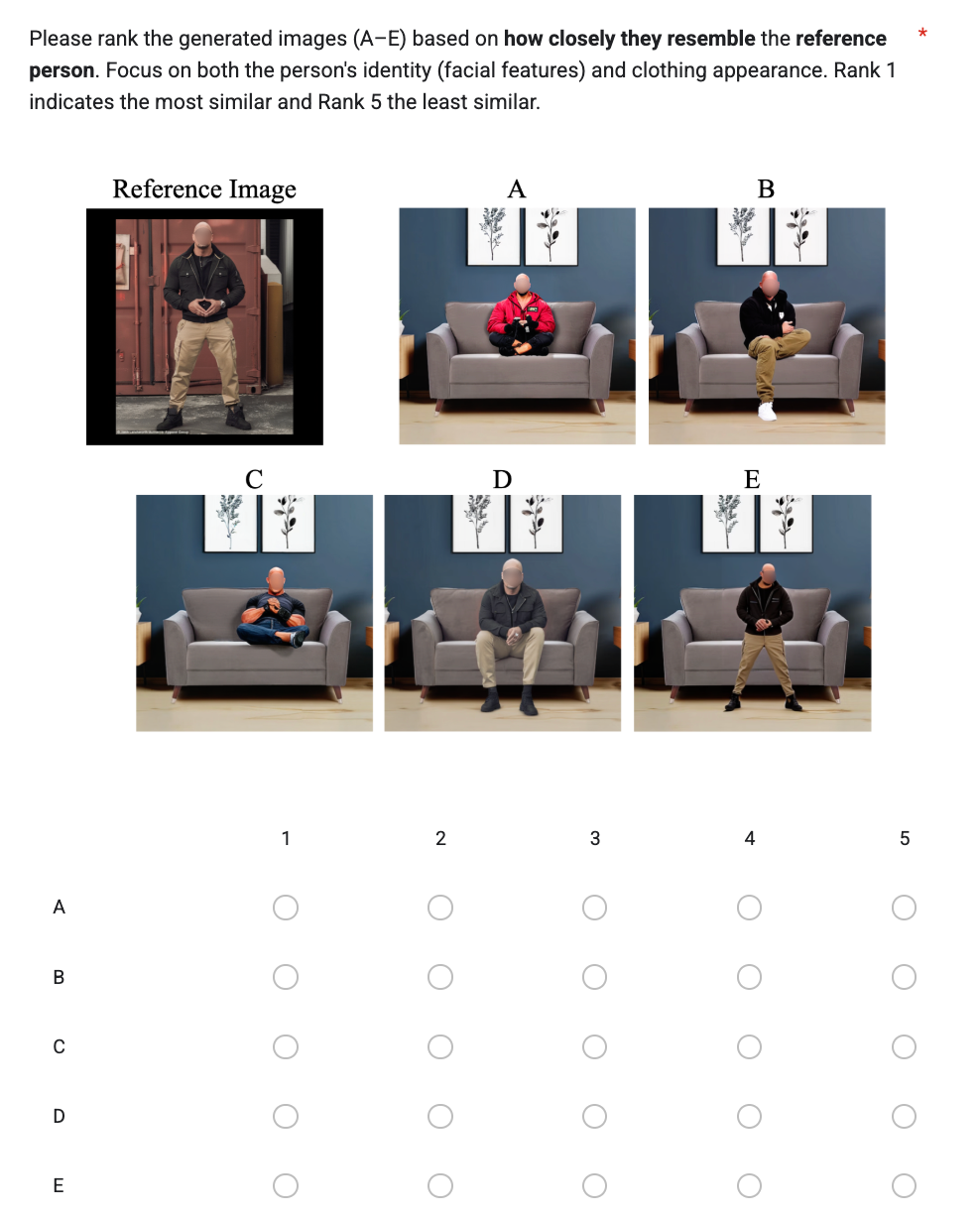}
    \caption{\textbf{Example questionnaire for evaluating subject fidelity.}}
    \label{fig: human-subject}
\end{figure*}

\begin{figure*}[h]
    \centering
    \includegraphics[width=0.75\linewidth]{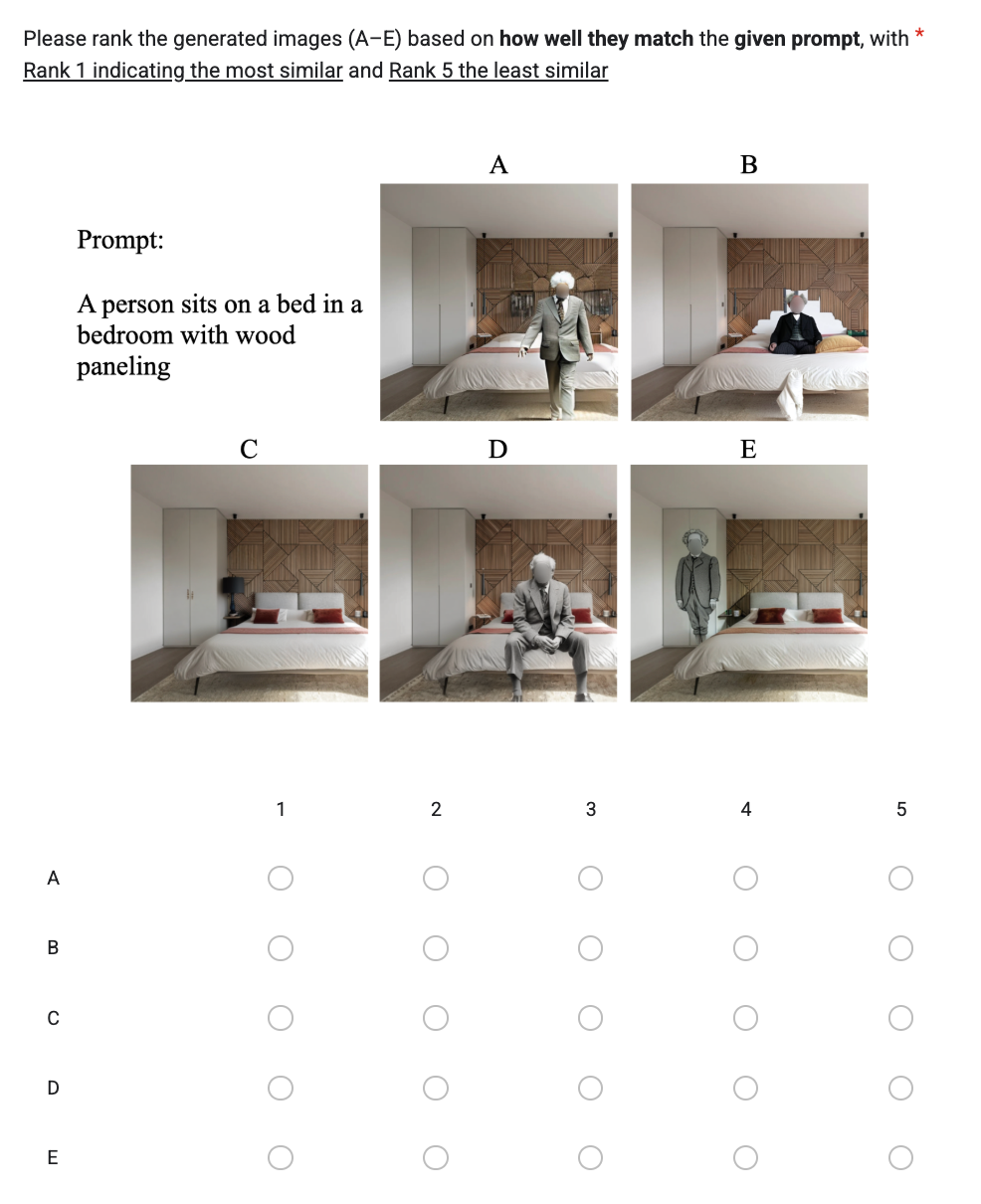}
    \caption{\textbf{Example questionnaire for evaluating prompt alignment.}}
    \label{fig: human-text}
\end{figure*}

\begin{figure*}[h]
    \centering
    \includegraphics[width=0.75\linewidth]{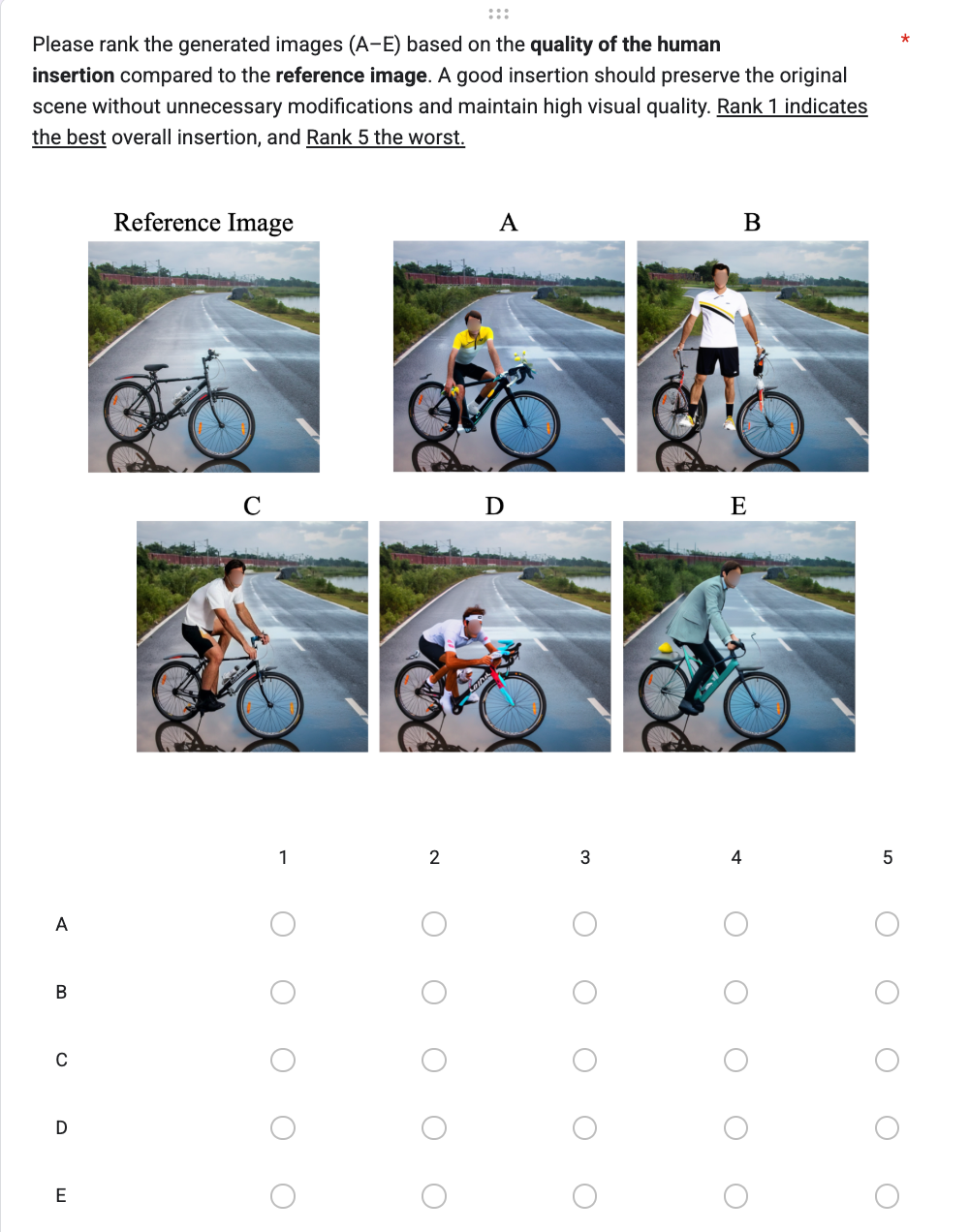}
    \caption{\textbf{Example questionnaire for evaluating background fidelity.}}
    \label{fig: human-background}
\end{figure*}

We provide example questions of the user study. For subject
fidelity, participants were presented with a reference image and several generated images using different methods, and were
asked to rank the generated images according to which better represents the subject in the reference image, as shown in Fig.~\ref{fig: human-subject}. For prompt alignment, the subjects were presented with the generated images alongside the text prompt used to generate these images, and were asked to rank the images according to which aligns best with the given prompt, as shown in Fig.~\ref{fig: human-text}. For background fidelity, the subjects were presented with the generated images alongs with the original scene image, and were asked to rank the images according to which aligns best with the original scene image, with an example shown in Fig.~\ref{fig: human-background}. A total number of 51 users responded to 36 ranking questions, resulting in a total of 1836 responses.

\newpage
\section{Implementation Details}
\subsection{Code Snippet}
\begin{figure*}[h]
    \centering
    \includegraphics[width=0.75\linewidth]{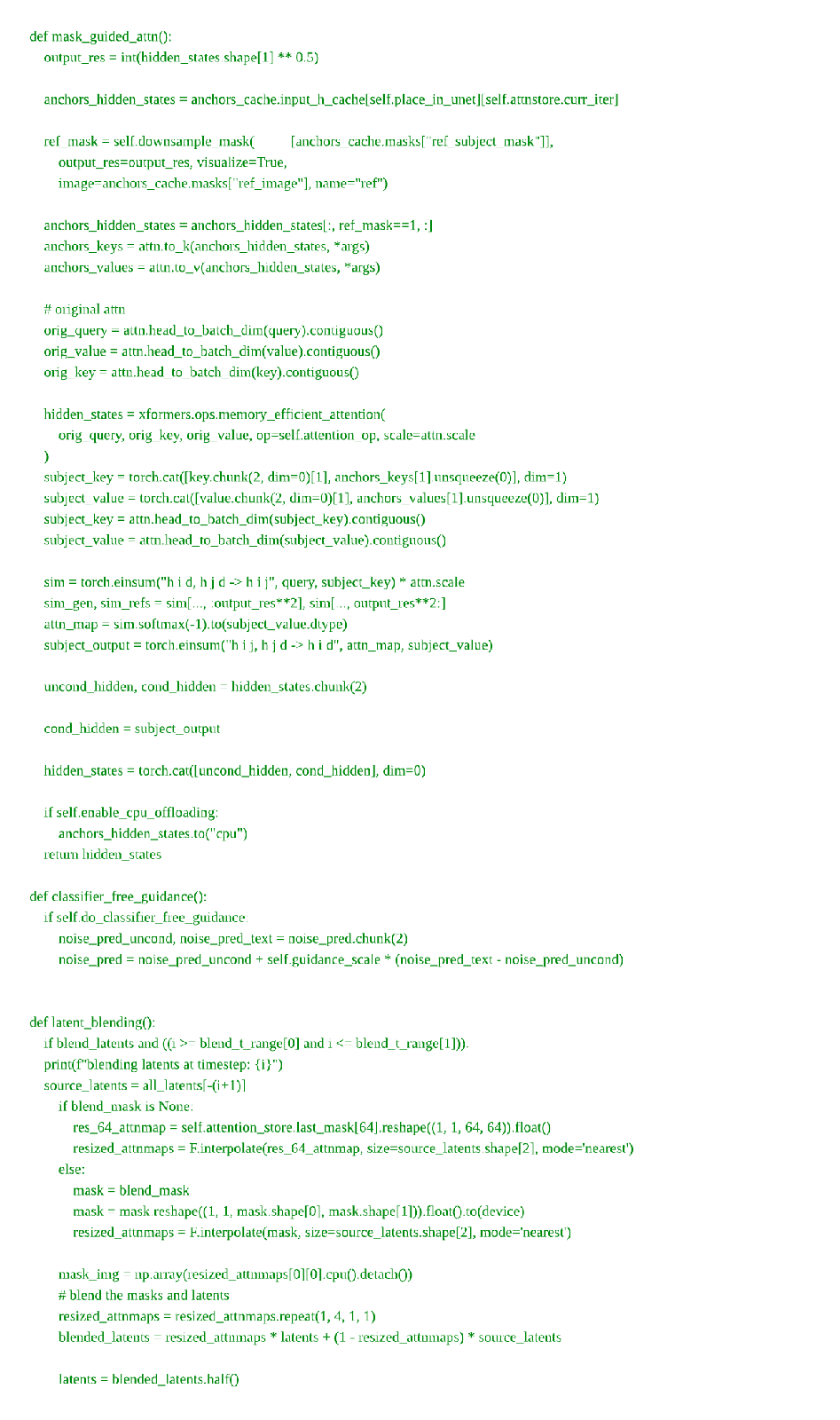}
    \caption{\textbf{Code snippets of the core components in \name.}}
    \label{fig:code}
\end{figure*}